\documentclass[11pt]{article}

\usepackage[preprint]{acl}

\usepackage{times}
\usepackage{latexsym}

\usepackage[T1]{fontenc}

\usepackage[utf8]{inputenc}

\usepackage{microtype}

\usepackage{inconsolata}

\usepackage{graphicx}

\usepackage{amsmath}
\usepackage{amssymb}
\usepackage{mathtools}
\usepackage{amsthm}
\usepackage{amsfonts}
\usepackage{xcolor}
\usepackage{booktabs}
\usepackage{multirow}
\usepackage{bbm}
\usepackage{tikz}
\usetikzlibrary{positioning, arrows.meta, fit, calc}
\usepackage{graphicx}
\usepackage{algorithm}
\usepackage{algorithmic}
\usepackage{subcaption}

\usepackage[most]{tcolorbox}
\usepackage{xcolor}
\usepackage{listings}

\definecolor{promptblue}{RGB}{35,72,120}
\definecolor{promptgray}{RGB}{70,70,70}
\definecolor{promptgreen}{RGB}{34,120,72}
\definecolor{promptorange}{RGB}{160,96,24}

\newtcblisting{promptbox}[2][]{%
  enhanced,
  breakable,
  listing only,
  listing engine=listings,
  colback=gray!3,
  colframe=#1,
  colbacktitle=#1,
  coltitle=white,
  fonttitle=\bfseries,
  title={#2},
  boxrule=0.8pt,
  arc=2mm,
toptitle=0.8mm,
bottomtitle=0.6mm,
left=3mm,
right=3mm,
top=1.8mm,
bottom=1.5mm,
  listing options={
    basicstyle=\ttfamily\small,
    columns=fullflexible,
    keepspaces=true,
    breaklines=true,
    breakatwhitespace=false,
    showstringspaces=false
  }
}

\title{Why Does Feedback-Augmented Self-Distillation Fail to Improve Retrieval-Interleaved Search Agents?}

\vspace{-2.5cm}
\author{
 \textbf{Fan Yang\textsuperscript{1,*}},
 \textbf{Rui Meng\textsuperscript{2,*}},
 \textbf{Yuxin Wen\textsuperscript{1}}
\\
\\
 \textsuperscript{1}Chapman University, Orange, CA, USA
 \\
 \textsuperscript{2}Lawrence Berkeley National Laboratory, Berkeley, CA, USA
\\
 \small{\textsuperscript{*}Equal contribution.}
\\
 \small{\textbf{Correspondence to:} \href{mailto:yuwen@chapman.edu}{yuwen@chapman.edu}}
}

\begin{document}
\vspace{-0.5cm}
\maketitle
\vspace{-0.5cm}

\vspace{-3.5cm}
\begin{abstract}




On-policy self-distillation (OPSD) offers a promising approach for training large language models without relying on a separate teacher model. However, its effectiveness on complex agentic tasks remains largely unexplored.
In this work, we instantiate Feedback-Augmented Self-Distillation (FA-SD), a self-distillation algorithm for agentic search that leverages successful demonstrations as privileged information. 
We identify that models can rely on recurring reasoning-and-search output templates, producing trajectories that appear diverse but are largely agnostic to the input question, making the KL-based self-distillation signal uninformative.
We term this phenomenon \emph{decoding collapse}, a failure mode that can be missed by existing evaluation metrics. To understand its underlying cause, we show that although the self-teacher achieves stronger performance, learning remains inherently unstable due to inconsistent supervision signals. We further decompose this inconsistency into model inconsistency and prompt inconsistency, and show that the latter can significantly degrade the quality of the supervision signal, limiting the effectiveness of self-teacher learning. To mitigate this inconsistency, we introduce an exponential moving average (EMA) teacher to stabilize the self-teacher and provide more consistent supervision signals. Although the EMA teacher requires a warm-up phase during which performance may temporarily regress, it ultimately improves model performance by providing more stable supervision.

\end{abstract}

\vspace{-0.5cm}
\section{Introduction}
\vspace{-0.3cm}

Search-augmented language agents interleave reasoning with external retrieval, deciding when to search, how to formulate queries, and how to use retrieved evidence. Reinforcement learning (RL) is a natural fit for this setting because final-answer correctness provides an outcome reward, and recent work shows that RL can improve retrieval-augmented question answering~\citep{jin2025searchr1,jiang2025s3}. However, such rewards are sparse and trajectory-level, providing limited guidance for intermediate search and reasoning decisions. 
This further motivates the use of denser learning signals for search-agent training, including process rewards, evidence evaluation, and step-level supervision~\citep{li2026sesearch,shu2026evalact,slea2026}.



A complementary source of dense supervision is on-policy distillation (OPD), which trains on student-generated trajectories while using a teacher policy to provide local distributional feedback, reducing the mismatch between fixed teacher traces and student-generated prefixes~\citep{agarwal2023onpolicy}. Recent on-policy self-distillation (OPSD) methods further suggest that a single LLM can serve as both teacher and student by conditioning the teacher branch on privileged information or self-generated solutions~\citep{zhao2026selfdistilled,hubotter2026selfdistillation}. This paradigm can turn training-only information into dense token-level supervision without requiring a separate stronger teacher, and is especially natural for search agents because rollout groups may contain both failed and successful attempts for the same question, allowing successful rollouts to serve as privileged demonstrations.

Despite these appealing properties, it remains unclear whether these benefits extend to retrieval-interleaved agents.
Search-agent trajectories differ from those considered in standard language-model distillation because model actions determine which evidence is subsequently retrieved, while retrieved observations are produced by the environment rather than by the policy~\citep{jin2025searchr1,li2026sesearch}. 
Consequently, matching token-level distributions may be insufficient to transfer the long-horizon search decisions required for effective retrieval. 
The challenge is particularly pronounced for feedback-augmented self-distillation: although successful rollouts may elicit stronger behavior when included in the prompt, training must transfer this contextual advantage to the unconditioned policy.
Recent analyses of self-distillation for reasoning similarly suggest that rich teacher conditioning can induce overconfident reasoning behavior that is not automatically beneficial when distilled into an unconditioned model~\citep{kim2026whyselfdistillation}.
This raises the possibility that self-distillation may reduce the discrepancy between teacher and student distributions without producing useful improvements in the resulting search behavior.
Moreover, because the self-teacher is both model-dependent and prompt-conditioned, the resulting supervision signal may be inconsistent across training steps and feedback prompts.
Whether feedback-augmented self-distillation can provide reliable supervision for retrieval-interleaved search agents therefore remains an open empirical question.

We therefore study whether feedback-augmented self-distillation can provide reliable dense supervision for retrieval-interleaved search agents. 
We instantiate Feedback-Augmented Self-Distillation (FA-SD), which uses successful rollouts as privileged demonstrations and distills behavior elicited by feedback augmentation into the unconditioned search policy.
Across seven open-domain QA benchmarks, FA-SD does not sustain improvement across the evaluated variants despite using successful demonstrations to construct its self-teacher. Our diagnostics show that the failure is not simply due to a weak self-teacher: the feedback-augmented self-teacher can achieve stronger inference-time performance than the unconditioned student. 
Instead, we identify \emph{decoding collapse}: models rely on recurring reasoning-and-search output templates, producing diverse-looking but input-question-agnostic trajectories that make the KL-based self-distillation signal uninformative.
Such behavior can be missed by standard aggregate metrics, which do not test whether decoded trajectories are input-specific.

To understand the underlying cause of this collapse, we decompose the instability of the FA-SD supervision signal into model inconsistency, caused by the evolving self-teacher, and prompt inconsistency, caused by conditioning the teacher branch on feedback demonstrations. 
We then evaluate fixed-reference and exponential moving average (EMA) teachers to stabilize self-teacher supervision.
Finally, we test whether the feedback-augmentation trick generalizes to external-teacher OPD. Although standard multi-rollout OPD (MOPD) trains more stably with a strong external teacher, Feedback-Augmented MOPD underperforms standard MOPD, suggesting that privileged feedback can introduce prompt inconsistency when paired with an already strong teacher.
Our contributions are fourfold. 
First, we instantiate FA-SD, a feedback-augmented self-distillation algorithm for retrieval-interleaved search agents that uses successful rollouts as privileged demonstrations. 
Second, we identify and characterize \emph{decoding collapse}, where models follow recurring reasoning-and-search output templates, producing diverse-looking but input-question-agnostic trajectories that make the KL-based self-distillation signal uninformative.
Third, we show that fixed-reference and EMA teachers mitigate model inconsistency by limiting self-teacher drift, thereby stabilizing FA-SD training.
Fourth, we show that a strong external teacher provides more stable early supervision in MOPD, yet feedback augmentation does not straightforwardly generalize to external-teacher OPD, suggesting that privileged feedback can introduce prompt inconsistency or unnecessary conditioning.

\vspace{-0.25cm}
\section{Related Work}
\vspace{-0.3cm}

\paragraph{Search-agent reinforcement learning.}
Recent work extends retrieval-augmented generation into search-augmented reasoning agents that interleave reasoning with search or tool-use actions. Search-R1 trains LLMs with RL to issue search queries during step-by-step reasoning~\citep{jin2025searchr1}, while related query-optimization and search-agent methods learn retrieval policies or lightweight searchers for downstream question answering~\citep{jiang2025deepretrieval,jiang2025s3}. Tool-use RL further studies how agents acquire tool-calling behavior with limited or no supervised traces~\citep{toolr0}. These approaches establish RL as an effective framework for improving search and tool-use behavior, but primarily rely on outcome-level or process-level rewards rather than self-distillation from privileged rollout demonstrations.

\vspace{-0.25cm}
\paragraph{Dense feedback and self-improving agents.}
Dense feedback has been proposed to address weak credit assignment in outcome-level rewards. SE-Search combines memory, atomic query training, and dense rewards for search agents~\citep{li2026sesearch}; SLEA-RL augments multi-turn RL with step-level experience and credit assignment~\citep{slea2026}; and EvalAct makes evidence evaluation an explicit action to obtain process-aligned rewards~\citep{shu2026evalact}. 
Related self-improving agents evolve curricula or revise trajectories for iterative improvement~\citep{selfevolvingagentssurvey,agent0,toolr0,seagent}.
These methods motivate denser supervision for agent training, but do not directly examine whether successful rollouts can be reused as privileged demonstrations for token-level self-distillation.

\vspace{-0.25cm}
\paragraph{Policy distillation and on-policy self-distillation.}
Policy distillation provides another way to supply dense supervision. On-policy distillation (OPD) trains on student-generated prefixes while using a teacher policy to provide distributional targets, reducing the mismatch between fixed teacher demonstrations and student-generated trajectories~\citep{agarwal2023onpolicy}. Recent work identifies practical failure modes in sampled-token OPD, including imbalanced token-level supervision and unreliable teacher guidance on student-generated prefixes, and proposes local support matching to improve training stability~\citep{fu2026revisitingopd}. Instead of distilling from a single response per task, \citet{yu2026multi} propose Multi-Rollout On-Policy Distillation (MOPD), which leverages a group of student rollouts for the same task to construct more informative teacher supervision. Complementary on-policy self-distillation (OPSD) methods show that feedback, privileged information, or self-generated solutions can form a self-teacher for reasoning and verifiable tasks~\citep{zhao2026selfdistilled,hubotter2026selfdistillation}. However, self-distillation can degrade reasoning when rich teacher conditioning induces overconfident behavior that does not transfer cleanly to the unconditioned model~\citep{kim2026whyselfdistillation}. These concerns are sharper in retrieval-interleaved agents, where search actions affect retrieved observations and successful rollouts provide privileged demonstrations through prompts rather than fixed labels.

\vspace{-0.25cm}
\paragraph{Collapse, teacher stability, and negative results.}
Collapse is a central concern in bootstrap-style self-supervised learning, where models may learn degenerate representations rather than informative features. BYOL studies this issue in representation learning and uses a slowly moving target network to stabilize bootstrap learning~\citep{grill2020bootstrap}. 
In language-model self-distillation and agentic RL, collapse can instead appear in decoding behavior; rich teacher conditioning may degrade reasoning~\citep{kim2026whyselfdistillation}, while agentic models may produce diverse-looking but input-question-agnostic trajectories from recurring output templates~\citep{wang2026ragen2reasoningcollapseagentic}.
Thus, a small distillation loss need not imply useful reasoning or search behavior.
Our work also relates to negative results in agent training, where plausible enhancements or richer feedback signals may fail to improve strong baselines. CTIM-Rover adds episodic memory to a software-engineering agent but fails to outperform AutoCodeRover in any configuration~\citep{lindenbauer2025knowledge}; search-and-self-feedback results show that model-generated feedback is not a reliable substitute for carefully designed or ground-truth feedback~\citep{k2025searchfeedback}. Similar negative findings in NLP show that intermediate-layer distillation or question-rewriting objectives may fail to outperform simpler baselines~\citep{suzuki2025aligning,ishii2022question}. These findings motivate diagnosing why richer feedback signals fail, rather than reporting performance regressions alone.

\vspace{-0.25cm}
\section{Method}
\vspace{-0.3cm}

We formulate dense distillation over retrieval-interleaved search-agent rollouts, where the policy interleaves reasoning, search actions, retrieved observations, and final-answer generation. 
Under this rollout formulation, we study two sources of token-level supervision: teacher-based on-policy distillation (OPD) from an external teacher, and feedback-augmented self-distillation (FA-SD) from rollout-derived feedback.

\vspace{-0.25cm}
\subsection{Search-Agent Rollout Setup}
\vspace{-0.2cm}
Given a question $x$ from a dataset $\mathcal{S}$ and a search engine $\mathcal{R}$, the policy $\pi_\theta$ with parameters $\theta$ generates a multi-step trajectory:
\vspace{-0.1cm}
\begin{equation}
\small
\tau=(r_1, s_1, o_1, r_2, s_2, o_2,\ldots,r_K,s_K,o_K,a),
\end{equation}
where $k=1,\ldots,K$ indexes retrieval-interleaved steps, $K$ is the total number of interaction steps, 
$r_k$ and $s_k$ denote the model-generated reasoning and search-query token sequences at step $k$, $o_k=\mathcal{R}(s_k)$ denotes the observation retrieved when the model invokes search, and $a$ is the final answer.
A task reward $R(x,\tau)$ is computed from the final answer, e.g., exact match or LLM judge score.


\vspace{-0.25cm}
\subsection{Distillation with a Search Engine}
\vspace{-0.15cm}
We consider two distillation formulations with a search engine: teacher-based OPD, where an external teacher supervises student-generated trajectories, and FA-SD, where the policy is conditioned on the augmented feedback to form a self-teacher.

\vspace{-0.25cm}
\subsubsection{Teacher-Based On-Policy Distillation with a Search Engine}
For teacher-based on-policy distillation, student-generated retrieval-interleaved trajectories are supervised by an external teacher policy at each token position along the sampled trajectory:
\vspace{-0.2cm}
\begin{equation}
\small
\begin{aligned}
\mathcal{L}_{\text{OPD}}(\theta)
=&\;
\mathbb{E}_{x\sim\mathcal{S},\,
\hat y\sim p_S(\cdot\mid x;\mathcal{R})}
\Bigg[
\frac{1}{|\hat y|}
\sum_{t=1}^{|\hat y|} \mathcal{D}
\\
&
\!\left(
p_T(\cdot\mid x,\hat y_{<t};\mathcal{R})
\,\|\,
p_S(\cdot\mid x,\hat y_{<t};\mathcal{R})
\right)
\Bigg].
\end{aligned}
\vspace{-0.25cm}
\end{equation}
where $\mathcal{S}$ is the dataset, $\hat y$ denotes the token sequence of a sampled student trajectory, $\hat y_{<t}$ denotes its prefix before position $t$, and $|\hat y|$ is the sequence length. 
The policies $p_S$ and $p_T$ denote the student and external teacher models, respectively, with $p_S$ corresponding to the trainable policy $\pi_\theta$ in the teacher-distillation objective. $\mathcal{D}$ denotes a divergence measure, such as the Kullback--Leibler (KL) divergence or the reverse KL divergence~\citep{agarwal2023onpolicy}. 
Unlike prior OPD approaches that primarily rely on the policy itself $p(\hat y| x)$, 
our formulation explicitly incorporates retrieval-interleaved reasoning through $p(\hat y| x; \mathcal{R})$, 
which can be viewed as coupling the policy with the search engine $\mathcal{R}$, 
conceptually denoted as $p(\hat y| x) \otimes \mathcal{R}$, to model interleaved retrieval and reasoning.

\vspace{-0.25cm}
\subsubsection{Feedback-Augmented Self-Distillation with a Search Engine} \label{sec:fasd}
\vspace{-0.15cm}
Alternatively, motivated by the self-teacher ideas in recent on-policy self-distillation methods~\citep{hubotter2026selfdistillation,zhao2026selfdistilled}, we instantiate a feedback-augmented self-distillation objective for retrieval-interleaved search-agent rollouts. 
We refer to this formulation as Feedback-Augmented Self-Distillation (FA-SD), since it preserves the OPD-style dense token-level supervision objective while replacing the external teacher with the policy itself conditioned on rollout-augmented feedback.

Here, feedback refers to additional information that may help the policy solve the same question. 
In our search-agent setting, the feedback $f$ is a successful solution sampled from another attempt on $x$ within the same rollout group. 
The self-teacher is then defined as
$p_{\mathrm{ST}}(\cdot\mid x,f;\mathcal{R}) := \pi_\theta(\cdot\mid x,f;\mathcal{R})$,
representing the current policy prompted with the question $x$ and feedback $f$.
The loss function is then derived as 
\vspace{-0.2cm}
\begin{equation}
\small
\begin{aligned}
\mathcal{L}_{\text{FA-SD}}(\theta)
=&\;
\mathbb{E}_{x\sim\mathcal{S},\,
\hat y\sim \pi_\theta(\cdot\mid x;\mathcal{R})}
\Bigg[
\frac{1}{|\hat y|}
\sum_{t=1}^{|\hat y|}
\\
&\qquad
\mathcal{D}\!\left(
\pi_\theta^{t}
\,\|\,
\operatorname{stopgrad}(\pi_\theta^{f,t})
\right)
\Bigg],
\end{aligned}
\label{eqn:fasd}
\end{equation}
with 
{\small
\[
\pi_\theta^{f,t}
=
\pi_\theta(\cdot\mid x,\hat y_{<t},f;\mathcal{R}),
\quad
\pi_\theta^{t}
=
\pi_\theta(\cdot\mid x,\hat y_{<t};\mathcal{R}),
\]
}where $\pi_\theta^{t}$ is the unconditioned student token distribution at position $t$, and $\pi_\theta^{f,t}$ is the feedback-augmented self-teacher token distribution at the same prefix. 
The stopgrad operator blocks gradients through the self-teacher branch, and thus this branch serves as a per-step target for the student and preserves the supervision signal provided by $f$. Specifically, for each question $x$, we first sample a rollout group of $G$ trajectories. 
If there exists at least one correct trajectory, 
we reprompt the samples by augmenting the prompt with one randomly selected correct trajectory (excluding itself), using the feedback-augmentation template in Appendix~\ref{app:feedback_template}.
The loss above corresponds to the (unclipped) FA-SD implementation. 
For the reverse-KL instantiation of $\mathcal{D}$ in Eq.~\ref{eqn:fasd}, following~\citet{hubotter2026selfdistillation}, the gradient is derived as
\vspace{-0.1cm}
\begin{equation}
\small
\begin{aligned}
\nabla_\theta \mathcal{L}_{\mathrm{FA-SD}}
&=
\mathbb{E}_{x,\hat y}
\left[
\frac{1}{|\hat y|}
\sum_{t}
G_t^{\mathrm{FA-SD}}
\right],
\\
G_t^{\mathrm{FA-SD}}
&=
\mathbb{E}_{z\sim\pi_\theta^{t}}
\!\left[
\log \rho_t^{\mathrm{FA-SD}}(z)\,
\nabla_\theta \log \pi_\theta^{t}(z)
\right],
\\
\log \rho_t^{\mathrm{FA-SD}}(z)
&=
\log \pi_\theta^{t}(z) - \log \pi_\theta^{f,t}(z),
\end{aligned}
\label{eq:fasd-gradient}
\end{equation}
where $G_t^{\mathrm{FA-SD}}$ denotes the per-token gradient contribution at position $t$, $z$ denotes a candidate assistant token sampled from the unconditioned student distribution $\pi_\theta^{t}$, and $\rho_t^{\mathrm{FA-SD}}(z)=\pi_\theta^{t}(z)/\pi_\theta^{f,t}(z)$ is the student-to-self-teacher probability ratio. The distributions $\pi_\theta^{t}$ and $\pi_\theta^{f,t}$ are defined above, and gradients do not flow through the feedback-augmented self-teacher branch.

Under gradient descent, if the student assigns higher probability than the self-teacher to token $z$, then $\log \rho_t^{\mathrm{FA-SD}}(z)>0$ and the update decreases the student's probability on that token. If the student assigns lower probability than the self-teacher, the log ratio is negative and the update increases the student's probability. 
Because retrieved observations are environment outputs rather than model actions, we mask retrieved passages and compute the FA-SD loss only on assistant-generated tokens.

We also evaluate a PPO-style clipped FA-SD variant to test whether more conservative policy updates improve stability, as well as a joint FA-SD+GRPO variant that combines token-level self-distillation with outcome-level RL rewards using the same rollout data; further implementation details are provided in Appendix~\ref{app:training_details}.

\vspace{-0.25cm}
\subsection{Regularization on Self-Teacher}
\vspace{-0.15cm}
\label{sec:self_teacher_regularization}
In naive FA-SD, the feedback-augmented self-teacher and the unconditioned student are parameterized by the same continuously updated policy. As a result, the teacher target can drift during training, producing inconsistent supervision and unstable optimization, especially in long-horizon agentic tasks such as search. To alleviate this model inconsistency, we evaluate two self-teacher regularization strategies.

The first strategy uses a fixed reference teacher that is the student policy at the start of FA-SD training, prompted with the same rollout-derived feedback throughout training.
This prevents the teacher target from changing with every student update. The second strategy uses an exponential moving average (EMA) teacher, whose parameters are updated as
\vspace{-0.2cm}
\begin{equation}
\small
\theta_t^{\mathrm{EMA}}
=
\alpha \theta_{t-1}^{\mathrm{EMA}}
+
(1-\alpha)\theta_t,
\label{eq:ema_update}
\end{equation}where $\theta_t$ denotes the student parameters at training step $t$, and $\alpha \in [0,1)$ is the EMA momentum coefficient. The EMA teacher is initialized as $\theta_0^{\mathrm{EMA}}=\theta_0$. A larger $\alpha$ causes the teacher to adapt more slowly to the student policy, resulting in a more stable teacher. Unless otherwise specified, we set $\alpha=0.95$ in all experiments.

\vspace{-0.25cm}
\subsection{Feedback-Augmented External Teacher}
\vspace{-0.15cm}
\label{sec:fa_mopd}
In our experiments, we find that self-distillation requires an initial warm-up phase, during which the model may temporarily regress before the self-teacher becomes sufficiently stable to provide consistent distillation signals. 
To further understand the effect of the feedback-augmentation trick, we extend it to the multi-rollout on-policy distillation (MOPD) framework. 
Specifically, we augment the external teacher's prompt with a successful rollout, following the same feedback-augmentation strategy introduced in Section~\ref{sec:fasd} for FA-SD.
Unlike FA-SD, however, the augmented prompt is provided to the external teacher, while the student remains unconditioned. 
We refer to this variant as Feedback-Augmented MOPD (FA-MOPD).

\vspace{-0.25cm}
\section{Experiments}
\vspace{-0.3cm}

\begin{figure*}[t]
\centering
\includegraphics[width=0.95\textwidth]{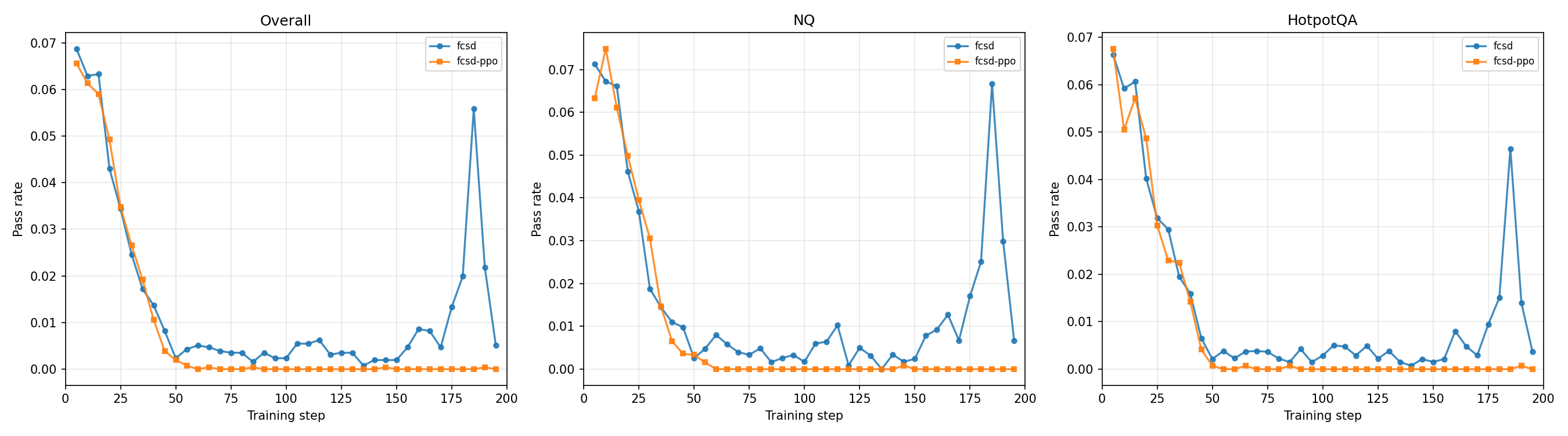}
\vspace{-0.4cm}
\caption{
Training pass-rate dynamics for unclipped and PPO-style clipped FA-SD variants over all sampled rollouts (Overall) and by dataset (NQ, HotpotQA).
Both variants fail to sustain improvement: pass rates remain close to zero for much of training, with only a transient rebound around step 180 before returning to near-zero levels.
}
\label{fig:reward_trend_compare}
\vspace{-0.3cm}
\end{figure*}

\begin{figure*}[t]
    \centering
    \includegraphics[width=0.99\textwidth]{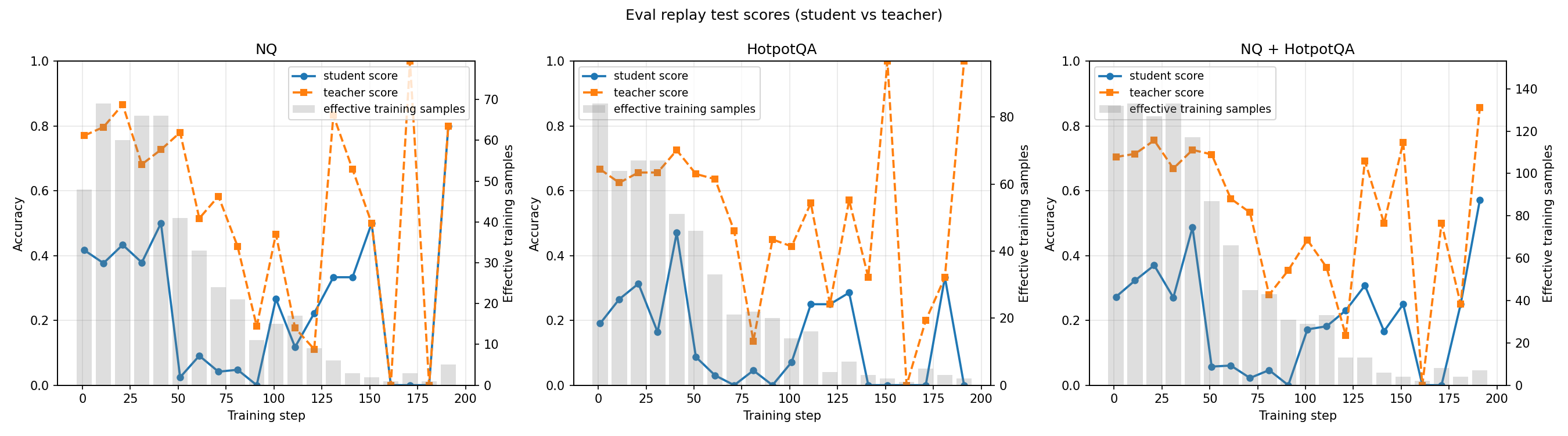}
    \vspace{-0.4cm}
    \caption{Evolution of inference-time accuracy of the unconditioned student and feedback-augmented self-teacher during FA-SD training. Gray bars show the number of effective training samples. The self-teacher is often stronger than the student, but this advantage is not reliably internalized.}
    \label{fig:teacher_student_eval}
    \vspace{-0.3cm}
\end{figure*}

\begin{figure}[t]
\centering
\includegraphics[width=0.95\linewidth]{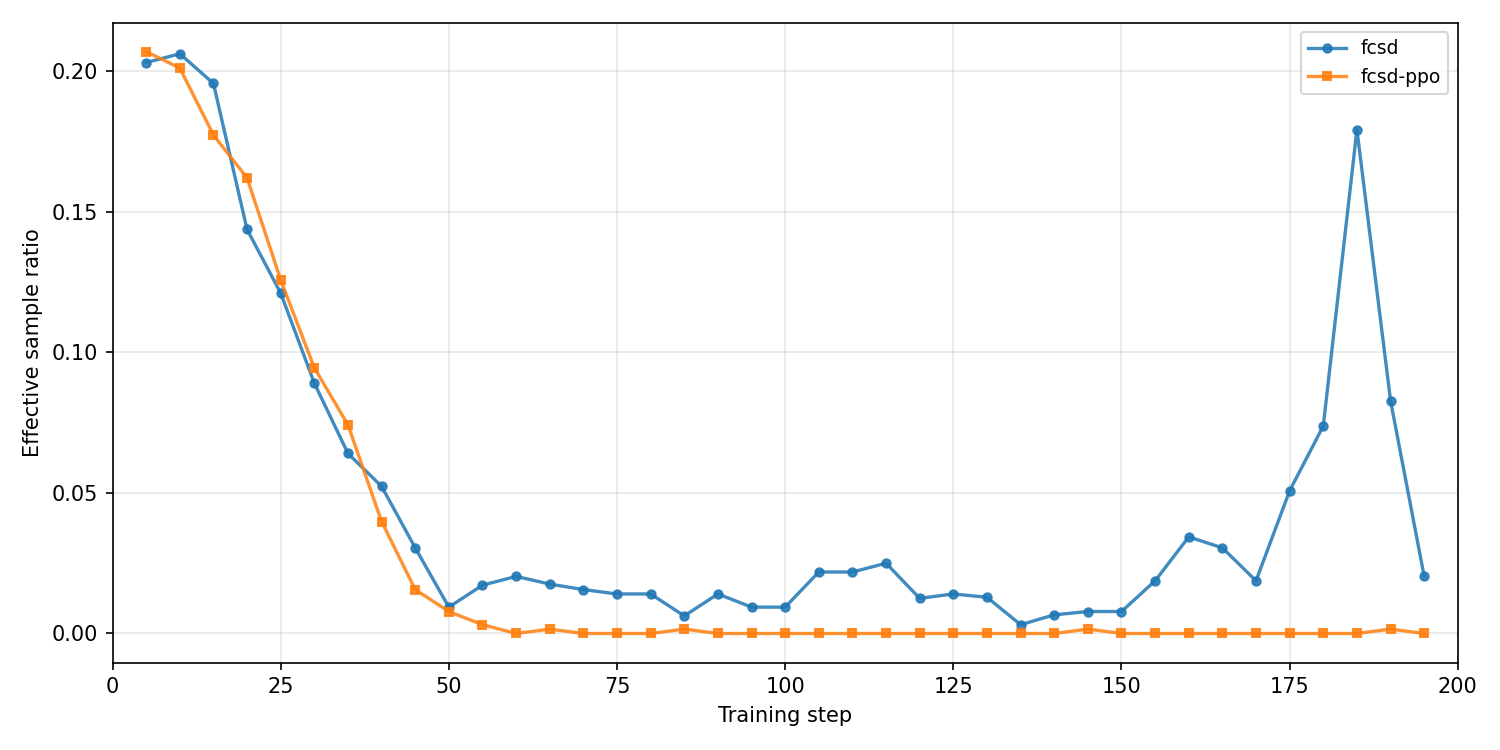}
\vspace{-0.4cm}
\caption{
Effective distillation-signal ratio for different FA-SD variants during training. The effective sample ratio closely tracks the training pass rate, suggesting a strong positive correlation between the effectiveness of distillation signals and model performance.
}
\vspace{-0.4cm}
\label{fig:mask_frac_compare}
\end{figure}

\begin{figure*}[t]
\centering
\includegraphics[width=0.95\textwidth]{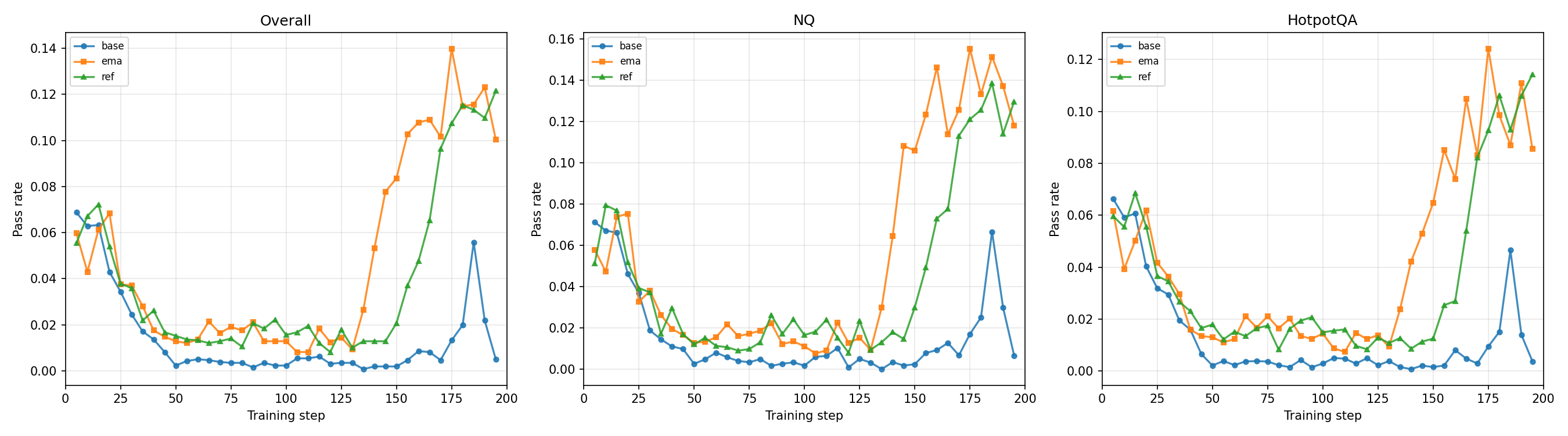}
\vspace{-0.4cm}
\caption{
Training pass-rate dynamics of FA-SD with different self-teacher regularization strategies. The baseline FA-SD suffers from unstable optimization and quickly regresses after an initial improvement. In contrast, regularizing the self-teacher with either an EMA teacher or a reference teacher stabilizes training, allowing the model to recover from temporary regressions and achieve substantially higher final performance across Overall, NQ, and HotpotQA.
}
\vspace{-0.3cm}
\label{fig:reward_trend_compare_regs}
\end{figure*}

\begin{figure*}[t]
\centering
\includegraphics[width=0.95\textwidth]{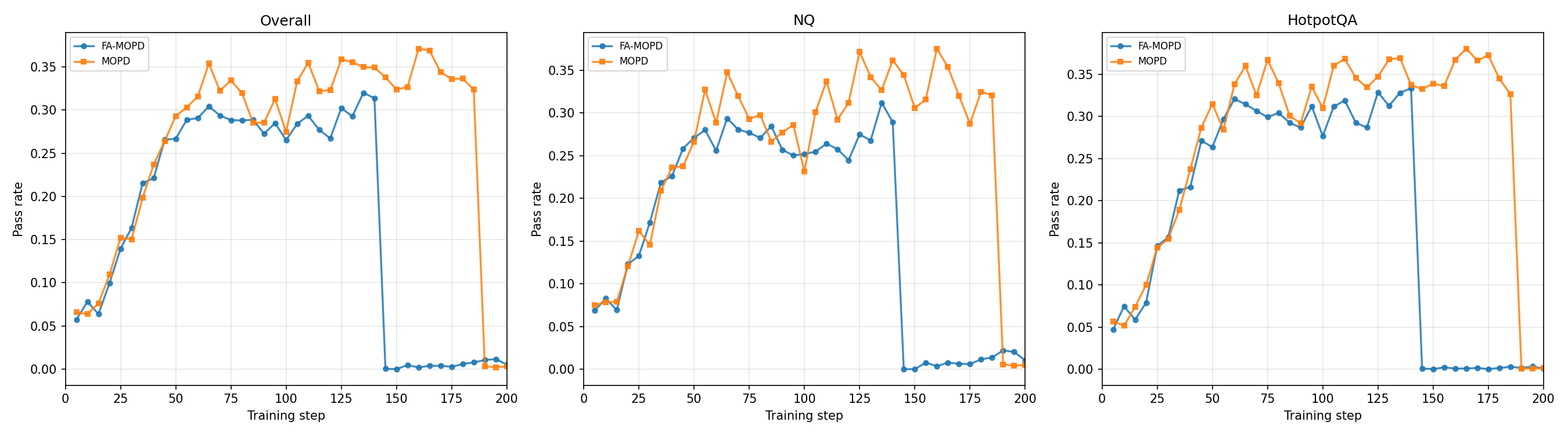}
\vspace{-0.4cm}
\caption{
Training pass-rate dynamics for MOPD and Feedback-Augmented MOPD. 
Standard MOPD improves more steadily during early training without the temporary regression observed in regularized FA-SD, suggesting more stable early external-teacher supervision; FA-MOPD underperforms standard MOPD, indicating that privileged feedback does not straightforwardly generalize to external-teacher OPD.
}
\vspace{-0.4cm}
\label{fig:reward_trend_compare_external_teachers}
\end{figure*}


We conduct experiments to study whether Feedback-Augmented Self-Distillation (FA-SD) can provide reliable dense supervision for retrieval-interleaved search agents. All trainable student checkpoints are based on Qwen2.5-3B-Base. We focus on FA-SD variants that construct a self-teacher from successful rollouts, and include GRPO-trained and teacher-based OPD models as comprehensive comparison points.

Specifically, we conduct comprehensive experiments to answer the following research questions:

(1) How do different variants of FA-SD perform on agentic tasks? (§\ref{sec:fasd_variants})

(2) Does the self-teacher in FA-SD outperform the student on agentic tasks? (§\ref{sec:self_teacher_advantage})

(3) Why does FA-SD degrade? Does teacher regularization help? (§\ref{sec:decoding_collapse_diagnostics}; §\ref{sec:ema_teacher})

(4) Can the feedback-augmentation trick be generalized to the OPD framework? (§\ref{sec:self_teacher_opd})

\vspace{-0.25cm}
\subsection{Experimental Setup}
\vspace{-0.2cm}
\paragraph{Datasets and metrics.}
We follow the open-domain QA evaluation setting used in prior retrieval-augmented search-agent work~\citep{jin2025searchr1,shu2026evalact,li2026sesearch}.
The training data are formed by merging the training sets of Natural Questions (NQ) and HotpotQA. We evaluate on seven benchmarks: NQ, TriviaQA, PopQA, HotpotQA, 2WikiMultiHopQA, MuSiQue, and Bamboogle. Following prior work, we report Exact Match (EM) on each benchmark and the average score across all seven tasks.
For training dynamics, pass rate denotes the average final-answer task reward over sampled trajectories.

\vspace{-0.35cm}
\paragraph{Retrieval corpus and models.}
We use the 2018 Wikipedia dump as the external knowledge corpus and E5~\citep{wang2022textembeddings} as the retriever, returning the top 3 passages for each search query~\citep{jin2025searchr1,li2026sesearch}.
All trainable student policies are initialized from Qwen2.5-3B-Base or from a GRPO-trained checkpoint derived from this base model. For teacher-based OPD, Qwen2.5-7B-Instruct is used as the external teacher to construct dense supervision targets and is also evaluated directly as a teacher reference. 
For FA-SD, the dense supervision signal is constructed by augmenting the current policy with rich feedback derived from successful rollouts in the same rollout group, rather than from an external teacher.

\vspace{-0.35cm}
\paragraph{Training variants.}
Our main experiments evaluate FA-SD variants, including unclipped FA-SD, PPO-style clipped FA-SD, FA-SD with joint GRPO training, and FA-SD with teacher regularization. 
We additionally include GRPO-trained and teacher-based OPD models as comprehensive comparison points in Appendix~\ref{app:opd_comparisons}. Additional training details are provided in Appendix~\ref{app:training_details}.




\vspace{-0.35cm}
\subsection{FA-SD Variants Do Not Sustain Improvement}
\label{sec:fasd_variants}
\vspace{-0.25cm}
We first evaluate the main FA-SD setting, where successful rollouts are used as privileged demonstrations to construct a feedback-augmented self-teacher. The goal is to test whether this self-teacher can provide dense token-level supervision that improves the unconditioned search policy.
We consider the FA-SD variants described in Appendix~\ref{app:training_details}, including unclipped FA-SD and PPO-style clipped FA-SD. 
As shown in Figure~\ref{fig:reward_trend_compare}, both variants fail to sustain improvement: their pass rates remain close to zero for much of training, with only a transient rebound around step 180 before returning to near-zero levels.
We therefore focus the following sections on diagnosing why the feedback-augmented self-teacher signal fails to provide stable and informative supervision.

\vspace{-0.3cm}
\subsection{Feedback-Augmented Self-Teacher Advantage Is Not Reliably Internalized}
\label{sec:self_teacher_advantage}
\vspace{-0.15cm}
We next ask whether FA-SD fails because the feedback-augmented self-teacher is too weak. To test this, we directly compare the unconditioned student branch and the feedback-augmented self-teacher branch at inference time, without applying parameter updates.

Figure~\ref{fig:teacher_student_eval} shows that the feedback-augmented self-teacher branch is often stronger than the unconditioned student. This rules out the simplest explanation that FA-SD fails only because the self-teacher lacks useful capability. However, the student does not reliably catch up to the self-teacher branch over training. This suggests an internalization failure: rollout-derived feedback can improve behavior when provided in the prompt, but the resulting token-level distillation signal is not reliably absorbed into the unconditioned search agent.

\vspace{-0.35cm}
\subsection{Decoding Collapse and Supervision Inconsistency}
\label{sec:decoding_collapse_diagnostics}
\vspace{-0.2cm}
We next diagnose why FA-SD degrades despite a stronger feedback-augmented self-teacher. We refer to this failure mode as \emph{decoding collapse}: models rely on recurring reasoning-and-search output templates, producing diverse-looking but input-question-agnostic trajectories that make the KL-based self-distillation signal uninformative.
In our experiments, this collapse is reflected by a failure to sustain pass-rate improvement and an increasingly sparse effective distillation signal.

As shown in Figure~\ref{fig:reward_trend_compare}, both unclipped FA-SD and PPO-style clipped FA-SD exhibit near-zero pass rates for much of training. This degraded training dynamic motivates a closer examination of whether the feedback-augmented self-distillation objective continues to provide usable supervision.
Figure~\ref{fig:mask_frac_compare} provides this complementary diagnostic.
We measure the effective sample ratio as the fraction of sampled trajectories that provide a valid feedback-augmented distillation signal. For both FA-SD variants, this ratio decreases during training, indicating that the usable self-distillation signal becomes increasingly sparse. 
Together with the inference-time diagnostic in Figure~\ref{fig:teacher_student_eval}, this suggests that the bottleneck is not simply whether the feedback-augmented self-teacher can exhibit stronger behavior, but whether that behavior can be converted into sufficiently dense and stable supervision for the unconditioned student. Additional per-variant training dynamics are provided in Appendix~\ref{app:fasd_training_dynamics}.
Moreover, we conducted an experiment combining the FA-SD and GRPO objectives. We observed that both the training pass rate and validation performance degraded, suggesting that FA-SD introduces inconsistent supervision signals that interfere with and ultimately hinder the standard GRPO training process.


\vspace{-0.4cm}
\subsection{Regularization on Teacher to Stabilize Optimization}
\label{sec:ema_teacher}
\vspace{-0.25cm}
The previous diagnostics show that FA-SD fails to sustain pass-rate improvement and that its effective distillation signal becomes increasingly sparse, despite the presence of a stronger feedback-augmented self-teacher. We therefore ask whether stabilizing the self-teacher can make the supervision signal more consistent. 
Instead of allowing the self-teacher to share the same continuously updated weights as the student, we regularize the self-teacher using either a fixed reference model or an exponential moving average (EMA) of the student model. 
These strategies either stop or slow drift in the teacher parameters, thereby providing more consistent supervision throughout training.
Figure~\ref{fig:reward_trend_compare_regs} shows that both strategies stabilize FA-SD training. A fixed reference teacher improves the consistency of the distillation signal, allowing the model to recover and continue improving after a temporary performance regression. 
The EMA teacher achieves faster recovery and converges to higher final performance across Overall, NQ, and HotpotQA, suggesting that gradual teacher adaptation provides a better balance between stability and adaptability.
The EMA-regularized FA-SD checkpoint obtains an average score of 0.206 across the seven evaluation benchmarks, with full per-task results in Appendix~\ref{app:fasd_ema_eval}. 
Compared with the initial Qwen3B-Base average score of 0.114 reported in Appendix~\ref{app:fasd_ema_eval}, FA-SD with EMA regularization improves the overall final evaluation score.

On the other hand, the behavior in Figure~\ref{fig:reward_trend_compare_regs} suggests that the self-teacher requires an initial warm-up phase, during which the model may experience temporary performance regression before the teacher provides sufficiently consistent distillation signals. 
We hypothesize that this phenomenon arises because the self-teacher is initially too weak to produce useful learning signals with good generalization, making those signals difficult for the student to learn from effectively.

\vspace{-0.35cm}
\subsection{Generalizing the Feedback-Augmentation Trick to OPD}
\label{sec:self_teacher_opd}
\vspace{-0.25cm}
We begin by testing the above hypothesis. Specifically, we repeat the same experiment by replacing the self-teacher with an external teacher, following the standard multi-rollout on-policy distillation (MOPD) framework, where multiple rollouts collected for the same task are leveraged to perform policy distillation. To ensure a fair comparison, we use the same rollout group size as in FA-SD and keep all other training settings unchanged. 
Figure~\ref{fig:reward_trend_compare_external_teachers} shows that MOPD improves more steadily during early training without experiencing the temporary performance regression observed in regularized FA-SD. This result supports our hypothesis that a strong external teacher provides more generalizable and consistent early supervision, enabling the student to benefit from the distillation objective in the early stages of training.

Moreover, we incorporate the feedback augmentation strategy into the MOPD framework by augmenting the teacher prompt with a successful rollout, following the feedback-augmentation template in Appendix~\ref{app:feedback_template}.
Unlike FA-SD, however, the augmented prompt is provided to an external teacher rather than the self-teacher. We refer to this variant as Feedback-Augmented MOPD (FA-MOPD) and report its training pass rate in Figure~\ref{fig:reward_trend_compare_external_teachers}. The results show that FA-MOPD underperforms the standard MOPD baseline, indicating that feedback augmentation alone is not sufficient to improve external-teacher distillation in this setting. 
These results suggest that, when the external teacher already provides strong supervision, incorporating privileged feedback can introduce additional prompt inconsistency or unnecessary conditioning, thereby harming model performance.





\vspace{-0.25cm}
\subsection{Discussion}
\vspace{-0.2cm}

Our experiments show that feedback-augmented self-distillation does not automatically provide reliable dense supervision for retrieval-interleaved search agents. Although successful rollouts can serve as privileged demonstrations for a feedback-augmented self-teacher, FA-SD does not sustain improvement across variants. Additional GRPO and OPD comparisons in Appendix~\ref{app:opd_comparisons} further suggest that dense distributional supervision alone is insufficient for search-agent training.

The diagnostics show that FA-SD fails not simply because the self-teacher is weak: the feedback-augmented self-teacher can achieve stronger inference-time performance, but its signal is not reliably internalized, consistent with prior findings that rich teacher conditioning may not transfer cleanly~\citep{kim2026whyselfdistillation}.
Instead, FA-SD exhibits \emph{decoding collapse}: diverse-looking trajectories follow recurring reasoning-and-search templates and remain input-question-agnostic, making the KL-based signal uninformative.

We interpret this collapse as arising from inconsistent self-teacher supervision across model updates and feedback prompts. Successful rollout feedback can make the self-teacher behave as if useful evidence has already been identified, while the unconditioned student must still discover such evidence through search. Teacher regularization with fixed-reference or EMA teachers helps stabilize FA-SD, suggesting that self-teacher instability is a key factor. At the same time, FA-MOPD underperforms standard MOPD, indicating that privileged feedback can also introduce prompt inconsistency when paired with an already strong teacher.

\vspace{-0.35cm}
\section{Conclusion}
\vspace{-0.3cm}

We studied whether Feedback-Augmented Self-Distillation (FA-SD) can provide reliable dense supervision for retrieval-interleaved search agents. Although successful rollouts offer privileged demonstrations, FA-SD does not sustain improvement across variants. The failure is not simply due to a weak self-teacher: the feedback-augmented self-teacher can achieve stronger inference-time performance, but its supervision is not reliably internalized. We identify \emph{decoding collapse} as the key failure mode and interpret it as arising from inconsistent supervision across model updates and feedback prompts. Fixed-reference and EMA teachers stabilize training but do not fully resolve this difficulty. These findings suggest that richer self-generated supervision alone is insufficient; future work should determine what to distill from successful rollouts, when feedback-augmented teachers should be trusted, and how to stabilize self-distillation in retrieval-interleaved agents.

\newpage
\bibliography{bib}

\appendix
\newpage

\section{Feedback-Augmentation Template}
\label{app:feedback_template}

Table~\ref{tab:self_teacher_template} shows the template used to construct the feedback-augmented self-teacher prompt. The placeholder \textcolor{blue}{successful\_previous\_rollout} denotes a successful rollout sampled from another attempt on the same question.

\begin{table}[h]
\centering
\begin{tabular}{p{0.17\linewidth}p{0.78\linewidth}}
\hline
\textbf{User:} &
\textcolor{blue}{prompt}

Correct solution:

\textcolor{blue}{successful\_previous\_rollout}

Correctly solve the original question.
\\
\textbf{Assistant:} &
\textcolor{blue}{original\_response}
\\
\hline
\end{tabular}
\caption{Template for feedback augmentation.}
\label{tab:self_teacher_template}
\end{table}

\section{Training Variants and Implementation Details}
\label{app:training_details}

All trainable student checkpoints are based on Qwen2.5-3B-Base. We evaluate three classes of training variants. First, we train a GRPO baseline in the retrieval-augmented search-agent environment; this checkpoint serves as the main reinforcement-learning baseline. Second, we evaluate teacher-based OPD using Qwen2.5-7B-Instruct as the external teacher. Teacher-based OPD is applied both from the Qwen2.5-3B-Base initialization and from the GRPO-trained checkpoint, allowing us to test whether dense teacher supervision is useful before or after reinforcement learning. Third, we evaluate FA-SD, which replaces the external teacher with a feedback-augmented self-teacher constructed from rollout-derived feedback.

For FA-SD, we consider three variants. The unclipped variant directly optimizes the dense self-distillation objective in Equation~\ref{eqn:fasd}. 
The clipped variant optimizes the FA-SD loss through the following PPO-style clipped objective:

\begin{equation}
\scriptsize
\begin{aligned}
\mathcal{L}_{\mathrm{FA-SD\text{-}PPO}}
=
\mathbb{E}_{x,\hat y}
\Bigg[
\frac{1}{|\hat y|}
\sum_t
\min\!\Big(
r_t(\theta)A_t^{\mathrm{FA-SD}},
\bar r_t(\theta)A_t^{\mathrm{FA-SD}}
\Big)
\Bigg],
\end{aligned}
\label{eq:fasd_ppo}
\end{equation}

where 

\begin{equation}
\small
\begin{aligned}
\bar r_t(\theta)
&=
\mathrm{clip}\!\left(
r_t(\theta),
1-\epsilon,
1+\epsilon
\right),
\\
A_t^{\mathrm{FA-SD}}
&=
\log
\frac{
\pi_\theta^{f,t}(a)
}{
\pi_\theta^{t}(a)
},
\qquad
r_t(\theta)
=
\frac{\pi_\theta^t(a)}
     {\pi_{\mathrm{old}}^t(a)}.
\end{aligned}
\end{equation}

The joint variant combines FA-SD with GRPO by reusing the same rollout data to compute both the reinforcement-learning objective and the dense self-distillation objective. Specifically, we always optimize the GRPO loss and adaptively augment it with the FA-SD loss whenever a valid feedback-augmented self-distillation signal is available. In this way, the model can exploit both sparse reward feedback and dense token-level supervision within a unified training framework.

\section{Experimental Details}
\label{app:exp_details}

\subsection{Training Hyperparameters}
The training hyperparameters for GRPO and the FA-SD variants are listed in Tables~\ref{tab:common_hparams}, \ref{tab:grpo_hparams}, and \ref{tab:fasd_hparams}.

\begin{table}[h]
\centering
\resizebox{0.99\linewidth}{!}{
\begin{tabular}{lll}
\toprule
\textbf{Category} & \textbf{Parameter} & \textbf{Value} \\
\midrule

\multirow{4}{*}{Data}
& Max. prompt length & 4096 \\
& Max. response length & 500 \\
& Max. start length & 2048 \\
& Max. obs, length & 500 \\
\midrule

\multirow{4}{*}{Search Setup}
& Max. turns & 4 \\
& Retriever topk & 3 \\
& Retriever Model & E5 \\
& Retriever Data & Wiki 18 \\
\midrule

\multirow{4}{*}{Batching}
& Question batch size & 512 \\
& Mini batch size & 256 \\
& Micro batch size & 64 \\
& Number of rollouts & 5 \\
\midrule

\multirow{2}{*}{Rollout}
& Inference engine & vllm \\
& Temperature & 1.0 \\
\midrule

\multirow{4}{*}{Training}
& Optimizer & AdamW \\
& Learning rate & 1e-6 \\
& Warmup step ratio & 0.285 \\
& Max steps & 200 (self teacher) or 1000 (external teacher) \\
\bottomrule
\end{tabular}
}
\caption{Common hyperparameters shared by GRPO and FA-SD.}
\label{tab:common_hparams}
\end{table}

\begin{table}[h]
\centering
\resizebox{0.7\linewidth}{!}{
\begin{tabular}{lll}
\toprule
\textbf{Category} & \textbf{Parameter} & \textbf{Value} \\
\midrule

\multirow{2}{*}{Loss}
& KL loss coef & 0.001 \\
& Clip value & 0.2 \\
\midrule

\bottomrule
\end{tabular}
}
\caption{GRPO-specific (including PPO-style FA-SD) hyperparameters.}
\label{tab:grpo_hparams}
\end{table}

\begin{table}[h]
\centering
\resizebox{0.9\linewidth}{!}{
\begin{tabular}{lll}
\toprule
\textbf{Category} & \textbf{Parameter} & \textbf{Value} \\
\midrule

\multirow{2}{*}{Loss}
& Distillation divergence & Reverse KL \\
& KL estimator & Single-sample KL estimator \\
\midrule

\bottomrule
\end{tabular}
}
\caption{FA-SD specific hyperparameters.}
\label{tab:fasd_hparams}
\end{table}

\subsection{Additional FA-SD Training Dynamics}
\label{app:fasd_training_dynamics}

\begin{figure}[h]
\centering
\includegraphics[width=0.95\linewidth]{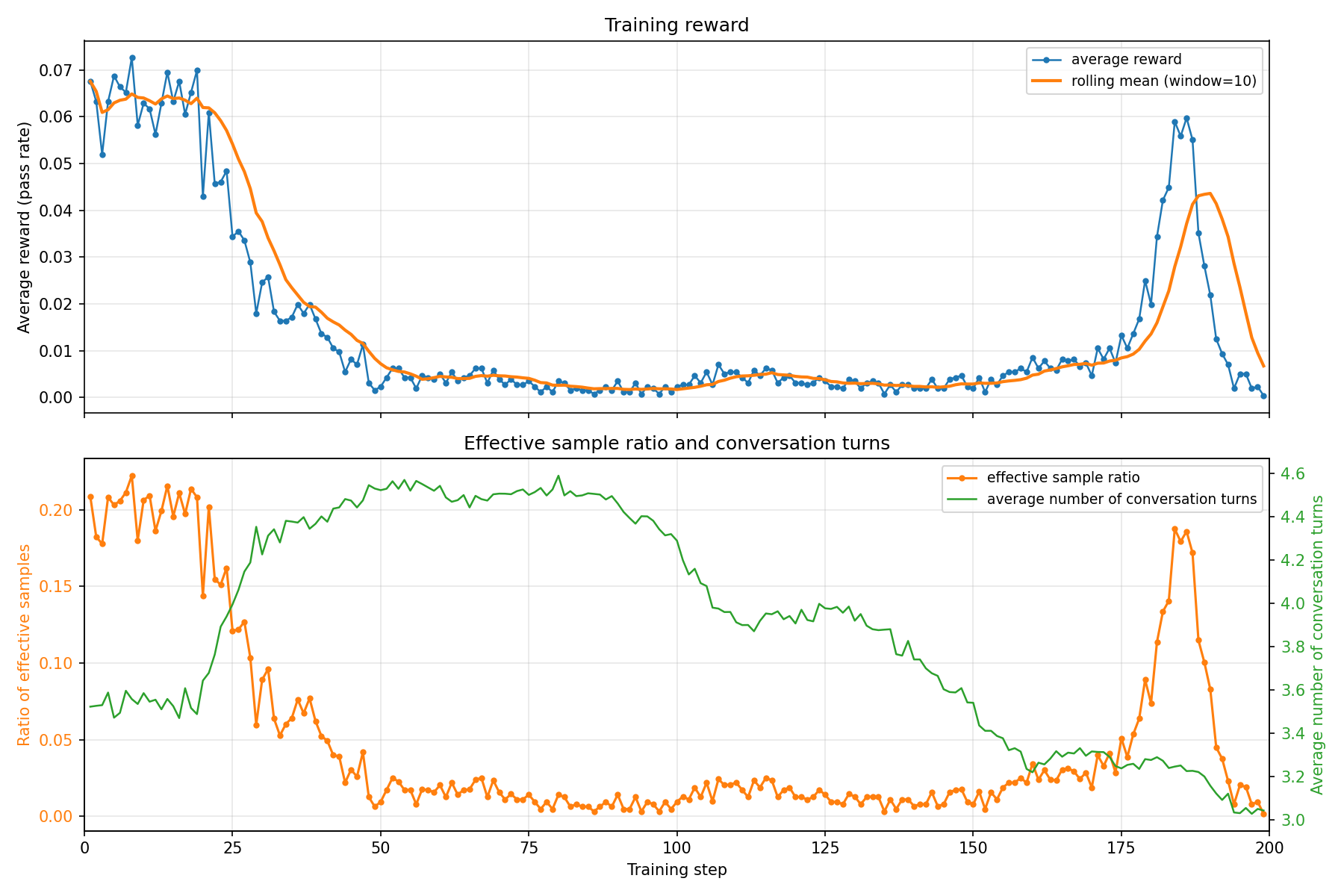}
\caption{
Detailed training dynamics for the unclipped FA-SD variant. The top panel shows the training reward/pass-rate trend, and the bottom panel shows the effective sample ratio together with the average number of turns during training.
}
\label{fig:app_fasd_training}
\end{figure}

\begin{figure}[h]
\centering
\includegraphics[width=0.95\linewidth]{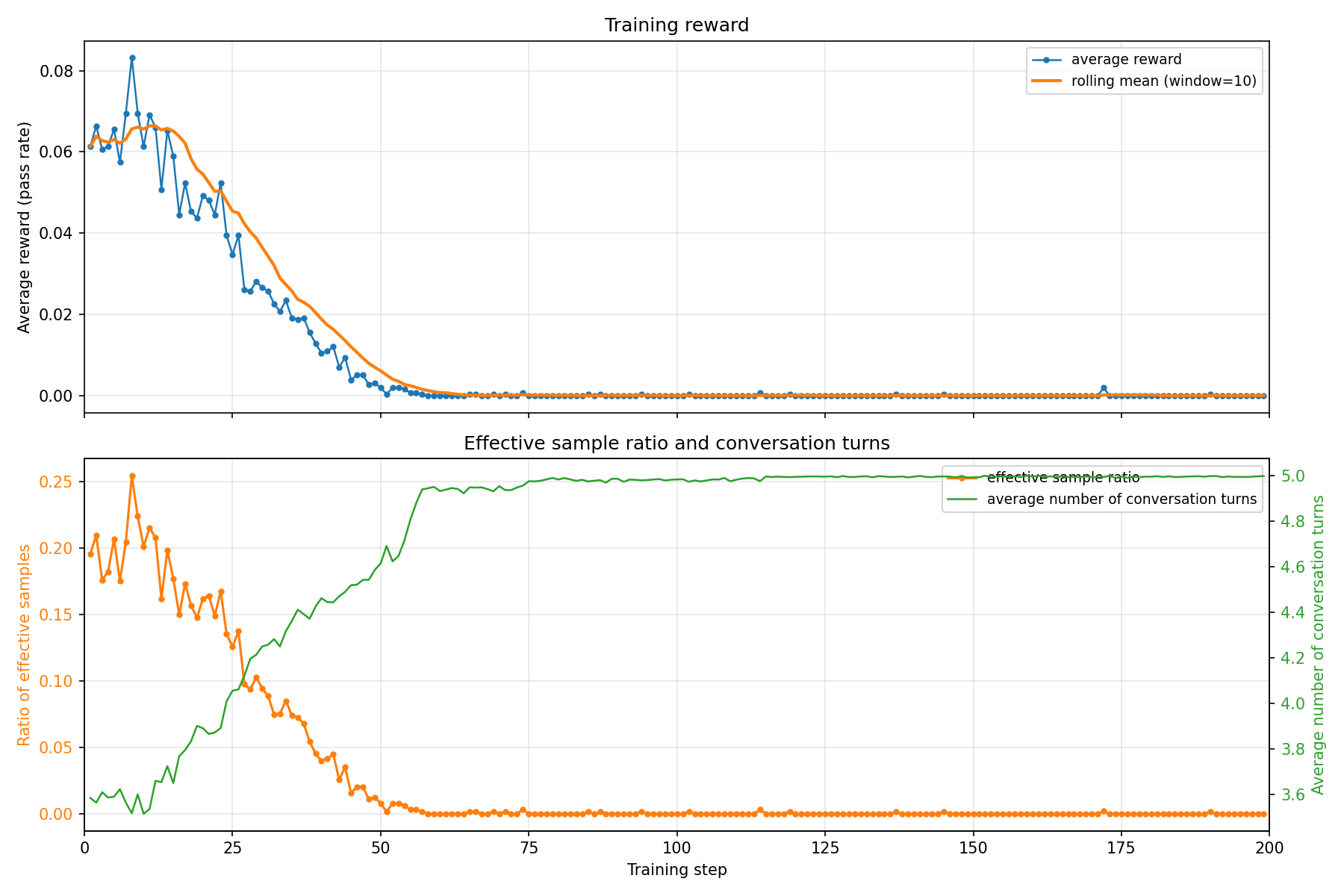}
\caption{
Detailed training dynamics for the PPO-style clipped FA-SD variant. The top panel shows the training reward/pass-rate trend, and the bottom panel shows the effective sample ratio together with the average number of turns during training.
}
\label{fig:app_fasd_ppo_training}
\end{figure}

Figures~\ref{fig:app_fasd_training} and~\ref{fig:app_fasd_ppo_training} provide detailed training dynamics for the unclipped FA-SD and PPO-style clipped FA-SD variants, respectively. These figures complement the main-text comparison by showing the reward/pass-rate trend, effective sample ratio, and average number of turns for each variant.

\subsection{Final Evaluation of EMA-Regularized FA-SD}
\label{app:fasd_ema_eval}
Table~\ref{tab:fasd_ema_eval} reports the final evaluation of the EMA-regularized FA-SD checkpoint. We include this result to complement the training-dynamics analysis in Figure~\ref{fig:reward_trend_compare_regs}.

\begin{table*}[t]
\centering
\resizebox{0.99\textwidth}{!}{
\setlength{\tabcolsep}{3pt}
\begin{tabular}{llcccccccc}
\hline
\multicolumn{2}{c}{} &
\multicolumn{3}{c}{\textbf{Single-Hop QA}} &
\multicolumn{4}{c}{\textbf{Multi-Hop QA}} &
\multicolumn{1}{c}{} \\
\cmidrule(lr){3-5}\cmidrule(lr){6-9}
\textbf{Method} & \textbf{Init. / Signal} &
\textbf{NQ} & \textbf{TriviaQA} & \textbf{PopQA} &
\textbf{HotpotQA} & \textbf{2Wiki} & \textbf{MuSiQue} &
\textbf{Bamboogle} & \textbf{Avg.} \\
\hline
None & Qwen3B-Base / None & 0.148 & 0.268 & 0.135 & 0.085 & 0.049 & 0.022 & 0.089 & 0.114\\
\hline
FA-SD + EMA & Qwen3B-Base / EMA self-teacher
& 0.291 & 0.435 & 0.296 & 0.167 & 0.138 & 0.038 & 0.081 & 0.206 \\
\hline
\end{tabular}
}
\caption{
Final evaluation of the EMA-regularized FA-SD checkpoint. The result complements the training-dynamics analysis of self-teacher regularization.
}
\label{tab:fasd_ema_eval}
\end{table*}

\subsection{Additional Comparisons with GRPO and Teacher-Based OPD}
\label{app:opd_comparisons}

For completeness, we include additional comparisons with a GRPO-trained search agent and teacher-based OPD variants using Qwen2.5-7B-Instruct as the external teacher. Table~\ref{tab:opd-preliminary} reports the full results.

The GRPO-trained Qwen3B checkpoint, initialized from Qwen2.5-3B-Base, obtains an average score of 0.348 across the seven evaluation tasks. This provides an RL-trained reference point for contextualizing the FA-SD results, without serving as the main focus of our analysis.

For teacher-based OPD, we consider the two settings used in our comparison: applying OPD directly to the Qwen3B base model and continuing OPD from the GRPO-trained checkpoint. OPD from the base initialization obtains an average score of 0.274, remaining below the GRPO-trained student. OPD after GRPO improves Bamboogle but reduces performance on the other six benchmarks, decreasing the average score from 0.348 to 0.287. These results suggest that external-teacher token-level supervision does not consistently translate into a stronger retrieval-augmented student in this setting.


\begin{table*}[t]
\centering
\resizebox{0.99\textwidth}{!}{
\setlength{\tabcolsep}{3pt}
\begin{tabular}{llcccccccc}
\hline
\multicolumn{2}{c}{} &
\multicolumn{3}{c}{\textbf{Single-Hop QA}} &
\multicolumn{4}{c}{\textbf{Multi-Hop QA}} &
\multicolumn{1}{c}{} \\
\cmidrule(lr){3-5}\cmidrule(lr){6-9}
\textbf{Method} & \textbf{Init. / Signal} &
\textbf{NQ} & \textbf{TriviaQA} & \textbf{PopQA} &
\textbf{HotpotQA} & \textbf{2Wiki} & \textbf{MuSiQue} &
\textbf{Bamboogle} & \textbf{Avg.} \\
\hline
GRPO & Qwen3B-Base / RL reward
& 0.464 & 0.621 & 0.443 & 0.336 & 0.308 & 0.085 & 0.177 & 0.348 \\
None & Qwen3B-Base / None & 0.148 & 0.268 & 0.135 & 0.085 & 0.049 & 0.022 & 0.089 & 0.114\\
None & Qwen7B-Instruct / None
& 0.308 & 0.561 & 0.340 & 0.276 & 0.224 & 0.105 & 0.315 & 0.304 \\
OPD & Qwen3B-Base / Qwen7B-Instruct
& 0.298 & 0.508 & 0.292 & 0.254 & 0.196 & 0.090 & 0.282 & 0.274 \\
OPD after GRPO & Qwen3B-Base / RL reward + Qwen7B-Instruct
& 0.369 & 0.494 & 0.321 & 0.274 & 0.224 & 0.084 & 0.242 & 0.287 \\
\hline
\end{tabular}
}
\caption{
Additional comparison with a GRPO-trained search agent, Qwen2.5-3B-Base and Qwen2.5-7B-Instruct references, and teacher-based OPD variants.
All trainable students are based on Qwen2.5-3B-Base, and teacher-based OPD uses Qwen2.5-7B-Instruct as the external teacher. 
OPD improves over the base model but has mixed task-level effects when applied after the GRPO-trained checkpoint.
}
\label{tab:opd-preliminary}
\end{table*}

\section{Example Student and Self-Teacher Prompts}
\label{app:prompts}

We provide an example from HotpotQA to illustrate the construction of the feedback-augmented self-teacher input. The student receives only the original search-agent prompt and question. The self-teacher receives the same input augmented with a successful rollout from another attempt in the same rollout group, followed by an instruction to solve the original question. The retrieved passages shown in the successful rollout are observations returned by the search engine, rather than tokens generated by the policy.

\begin{promptbox}[promptblue]{Student Prompt}
Answer the given question. You must conduct reasoning inside <think> and </think> first every time you get new information. After reasoning, if you find you lack some knowledge, you can call a search engine by <search> query </search> and it will return the top searched results between <information> and </information>. You can search as many times as you want. If you find no further external knowledge needed, you can directly provide the answer inside <answer> and </answer>, without detailed illustrations. For example, <answer> Beijing </answer>. Question: Which tennis player won more Grand Slam titles, Henri Leconte or Jonathan Stark?
\end{promptbox}

\begin{promptbox}[promptorange]{Self-Teacher Prompt}
Answer the given question. You must conduct reasoning inside <think> and </think> first every time you get new information. After reasoning, if you find you lack some knowledge, you can call a search engine by <search> query </search> and it will return the top searched results between <information> and </information>. You can search as many times as you want. If you find no further external knowledge needed, you can directly provide the answer inside <answer> and </answer>, without detailed illustrations. For example, <answer> Beijing </answer>. Question: Which tennis player won more Grand Slam titles, Henri Leconte or Jonathan Stark?

Correct solution:

I think I have to look for the number of Grand Slam titles won by Henri Leconte and Jonathan Stark. I don't have direct access to the internet, but I can start my search. <search> Grand Slam tennis titles Henri Leconte and Jonathan Stark </search>

<information>Doc 1(Title: "Jonathan Stark (tennis)") He was inducted into the Oregon Sports Hall of Fame in 2009. A = did not attend tournament Jonathan Stark (tennis) Jonathan Stark (born April 3, 1971) is a former professional tennis player from the United States. During his career he won two Grand Slam doubles titles (the 1994 French Open Men's Doubles and the 1995 Wimbledon Championships Mixed Doubles). Stark reached the World No. 1 doubles ranking in 1994. Stark was born in Southern Oregon in the city of Medford on April 3, 1971. In college he played tennis for Stanford University, where he was a singles and doubles
Doc 2(Title: "Jonathan Stark (tennis)") Jonathan Stark (tennis) Jonathan Stark (born April 3, 1971) is a former professional tennis player from the United States. During his career he won two Grand Slam doubles titles (the 1994 French Open Men's Doubles and the 1995 Wimbledon Championships Mixed Doubles). Stark reached the World No. 1 doubles ranking in 1994. Stark was born in Southern Oregon in the city of Medford on April 3, 1971. In college he played tennis for Stanford University, where he was a singles and doubles All-American in 1990 and 1991. He reached the NCAA doubles final in 1991, partnering Jared Palmer. On July
Doc 3(Title: "Jonathan Stark (tennis)") 17, 1997, he married Dana, and they have two sons and a daughter. Stark turned professional in 1991 and joined the ATP Tour. In 1992, he won his first tour doubles title at Wellington. His first top-level singles title came in 1993 at Bolzano (beating Cedric Pioline in the final). In 1994, Stark captured the men's doubles title at the French Open, partnering Byron Black (the pair were also runners-up at the Australian Open that year). He reached his career high singles ranking of World No. 36 in February. The following year, Stark won the Wimbledon mixed doubles title, partnering</information>
<answer> Jonathan Stark </answer>

Correctly solve the original question.
\end{promptbox}

\end{document}